\documentclass[review]{elsarticle}

\usepackage{lineno,hyperref,subcaption}
\usepackage{algorithm}
\usepackage{amsmath}
\usepackage{amsfonts}
\usepackage{amssymb}
\usepackage[noend]{algpseudocode}
\usepackage{xcolor}
\modulolinenumbers[5]

\journal{}
\begin{document}

\begin{frontmatter}

\title{Magnetic-Visual Sensor Fusion based Medical SLAM for Endoscopic Capsule Robot}
%\tnotetext[mytitlenote]{Fully documented templates are available in the elsarticle package on \href{http://www.ctan.org/tex-archive/macros/latex/contrib/elsarticle}{CTAN}.}

%% Group authors per affiliation:
%\author{Elsevier\fnref{myfootnote}}
%\address{Radarweg 29, Amsterdam}
%\fntext[myfootnote]{Since 1880.}

%% or include affiliations in footnotes:
\author[address1,address2]{Mehmet Turan}
\ead{mturan@student.ethz.ch}

\author[address3]{Yasin Almalioglu}
\ead{yasin.almalioglu@boun.edu.tr}

\author[address5]{Hunter Gilbert}
\ead{hbgilbert@lsu.edu}

\author[address4]{Helder Araujo}
\ead{helder@isr.uc.pt}

\author[address2]{Ender Konukoglu}
\ead{ender.konukoglu@vision.ee.ethz.ch}

\author[address1]{Metin Sitti}
\ead{sitti@is.mpg.de}

%\author[mysecondaryaddress]{Global Customer Service\corref{mycorrespondingauthor}}
%\cortext[mycorrespondingauthor]{Corresponding author}
%\ead{support@elsevier.com}

\address[address1]{Max Planck Institute for Intelligent Systems, Stuttgart, Germany}
\address[address2]{Computer Vision Laboratory, ETH Zurich, Switzerland}
\address[address3]{Computer Engineering Department, Bogazici University, Turkey}
\address[address5]{Louisiana State University, Baton Rouge, LA, USA}
\address[address4]{Robotics Laboratory, University of Coimbra, Portugal}

\begin{abstract}
A reliable, real-time simultaneous localization and mapping (SLAM) method is crucial for the navigation of actively controlled capsule endoscopy robots.
These robots are an emerging, minimally invasive diagnostic and therapeutic technology for use in the gastrointestinal (GI) tract.
In this study, we propose a dense, non-rigidly deformable, and real-time map fusion approach for actively controlled endoscopic capsule robot applications.
The method combines magnetic and vision based localization, and makes use of frame-to-model fusion and model-to-model loop closure.
The performance of the method is demonstrated using an ex-vivo porcine stomach model. 
Across four trajectories of varying speed and complexity, and across three cameras, the root mean square localization errors range from 0.42 to 1.92 cm, and the root mean square surface reconstruction errors range from 1.23 to 2.39 cm.
\end{abstract}


\end{frontmatter}

%\linenumbers

\section{Introduction}
Gastrointestinal diseases are the primary diagnosis for about 28 million  patient visits per year in the United States\cite{NCHS}.
In many cases, endoscopy is an effective diagnostic and therapeutic tool, and as a result about 7 million upper and 11.5 million lower endoscopies are carried out each year in the U.S.\ \cite{peery2012burden}.  
Wireless capsule endoscopy (WCE), introduced in 2000 by Given Imaging Ltd., has revolutionized patient care by enabling inspection of regions of the GI tract that are inaccessible with traditional endoscopes, and also by reducing the pain associated with traditional endoscopy \cite{iddan2000wireless}. 
Going beyond passive inspection, researchers are striving to create capsules that perform active locomotion and intervention \cite{moglia2007wireless}.
With the integration of further functionalities, e.g. remote control, biopsy, and embedded therapeutic modules, WCE can become a key technology for GI diagnosis and treatment in near future. 

Several research groups have recently proposed active, remotely controllable robotic capsule endoscope prototypes equipped with additional operational functionalities, such as highly localized drug delivery, biopsy, and other medical functions \citep{sitti2015biomedical, yim2013magnetically, yim2012shape, yim2012design, goenka2014capsule, nakamura2008capsule, munoz2014review, carpi2011magnetically, keller2012method, mahoney2013managing, yim2014biopsy, petruska2013omnidirectional}. 
To facilitate effective navigation and intervention, the robot must be accurately localized and must also accurately perceive the surrouding tissues. 
Three-dimensional intraoperative SLAM algorithms will therefore be an indispensable component of future active capsule systems.

Several localization methods have been proposed for robotic capsule endoscopes such as fluoroscopy \citep{than2012review}, ultrasonic imaging \citep{yim20133}, positron emission tomography (PET) \citep{than2012review, yim20133}, magnetic resonance imaging (MRI) \citep{than2012review}, radio transmitter based techniques, and magnetic field-based techniques \citep{yim20133, son20165, popek2013localization, di2013real}. 
It has been proposed that combinations of sensors, such as {RF} range estimation and visual odometry, may improve the estimation accuracy \cite{geng2015accuracy, bao2015hybrid}. Morover, solutions that incorporate vision are attractive because a camera is already present on capsule endoscopes, and vision algorithms have been widely applied for robotic localization and map reconstruction. 
 
Feature-based SLAM methods have been applied on endoscopic type of image sequences in the past e.g \citep{mountney2009dynamic,grasa2014visual,stoyanov2010real, liu2009capsule}. As improvements to accomodate the flexibility of the GI tract, \citep{mountney2010motion} suggested a motion compensation model to deal with peristaltic motions, whereas \citep{mountney2006simultaneous} proposed a learning algorithm to deal with them. \citep{lin2013simultaneous} adapted parallel tracking and mapping techniques to a stereo-endoscope to obtain reconstructed 3D maps that were denser when compared to monoscopic camera methods.  \citep{mahmoud2016orbslam} has  applied ORB features to track the camera and proposed a method to densify the reconstructed 3D map, but pose estimation and map reconstruction are still not accurate enough.
All of these methods can fail to produce accurate results in cases of low-texture areas, motion blur, specular highlights, and sensor noise -- all of which are typically present during endoscopy. 
In this paper, we propose that a non-rigid and dense RGB Depth fusion method, which combines magnetic localization and visual pose estimation using particle filtering and recurrent neural network-based motion model estimation, can provide real-time, accurate localization and mapping for wireless capsule robots. 
We demonstrate the system in a pig stomach model by measuring the system performance in terms of both capsule localization and surface mapping accuracy.

\section{System Overview and Analysis}

\begin{figure}
\centering
% Use the relevant command to insert your figure file.
% For example, with the graphicx package use
  \includegraphics[width=\textwidth]{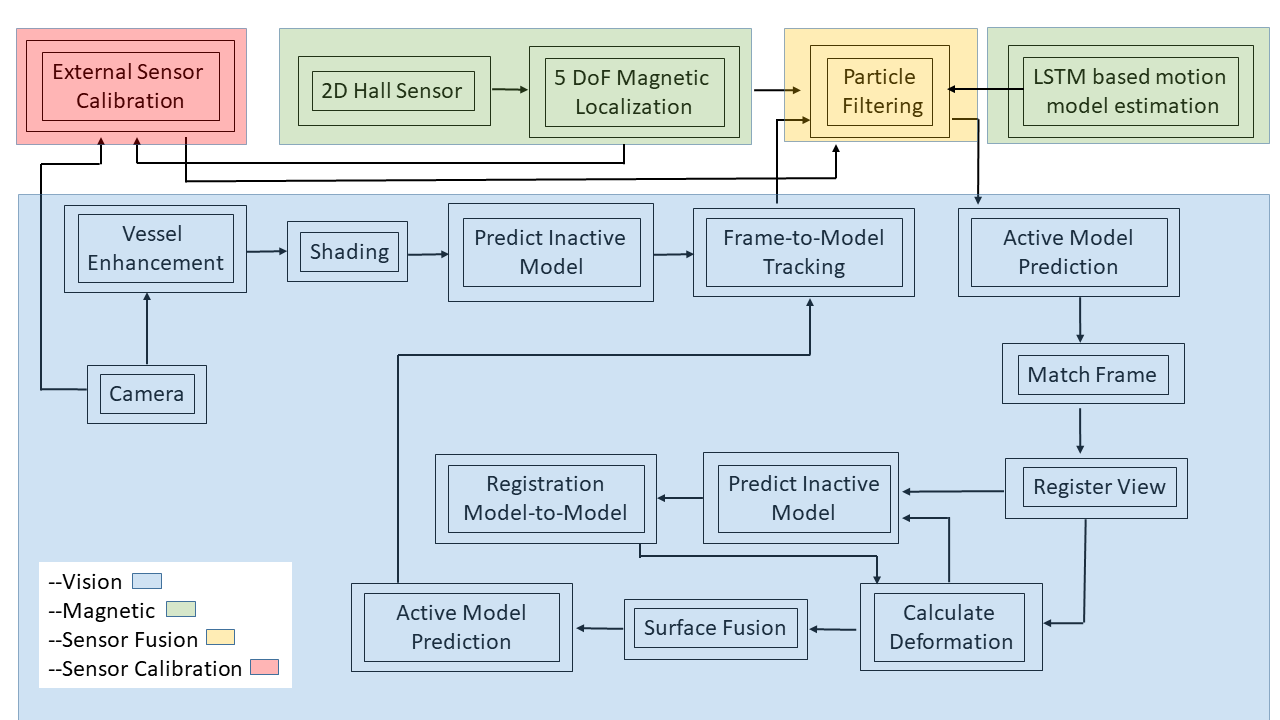}
% figure caption is below the figure
\caption{System overview}
\label{fig:model_flowchart}       % Give a unique label
\end{figure}

The system architecture of the method is depicted in Figure \ref{fig:model_flowchart}. Alternating between localization and mapping, our approach performs frame-to-model 3D map reconstruction in real-time. Below we summarize key steps of the proposed system:

\begin{itemize}
			
\item Perform offline inter-sensor calibration between magnetic hall sensor array and capsule camera system; 
			
\item Estimate endoscopic capsule robot pose in 5-DoF using magnetic localization system;

\item Estimate endoscopic capsule robot pose in 6-DoF using visual joint photometric-geometric frame-to-model pose optimization;

\item Fuse magnetic and visual pose estimations using back propagation based particle filtering and recurrent neural network based motion model estimation;

\item Perform non-rigid frame-to-model registration making use of particle filter based pose estimation and deformation constraints defined by the graph equations; 
\item To perform loop closure, in case there exists an intersection of the active model with inactive model within the current frame, fuse intersecting regions and deform the entire model non-rigidly.

\end{itemize}

\section{Magnetically Actuated Soft Capsule Endoscopes (MASCE)}
Our capsule prototype is a magnetically actuated soft capsule endoscope (MASCE) designed for disease detection, drug delivery and biopsy operations in the upper GI-tract. 
The prototype is composed of a RGB camera, a permanent magnet, an empty space for drug chamber and a biopsy tool (see Fig \ref{fig:exp_setup} for visual reference). 
The magnet exerts magnetic force and torque on the robot in response to a controlled external magnetic field. 
The magnetic torque and forces are used to actuate the capsule robot and to release drug. 
Magnetic fields from the electromagnets generate the magnetic force and torque on the magnet inside MASCE so that the robot moves inside the workspace. 
Sixty-four three-axis magnetic sensors are placed on the top, and nine electromagnets are placed in the bottom.

\section{Methods}

\subsection{Magnetic Localization System}
Our 5-DoF magnetic localization system is designed for the position and orientation estimation of untethered mesoscale magnetic robots \cite{son20165}.
The system uses an external magnetic sensor system and electromagnets for the localization of the magnetic capsule robot. 
A 2D-Hall-effect sensor array measures the component of the magnetic field from the permanent magnet inside the capsule robot at several locations outside of the robotic workspace. 
Additionally, a computer-controlled magnetic coil array consisting of nine electromagnets generates the magnetic field for actuation. 
The core idea of our localization technique is the separation of the capsule's magnetic field component from the actuator's magnetic field component. 
For that purpose, the actuator's magnetic field is subtracted from the magnetic field data which is acquired by a Hall-effect sensor array. 
As a further step, second-order directional differentiation is applied to reduce the localization error. 

\begin{figure}[t!]
	\centering
	\includegraphics[width=\columnwidth]{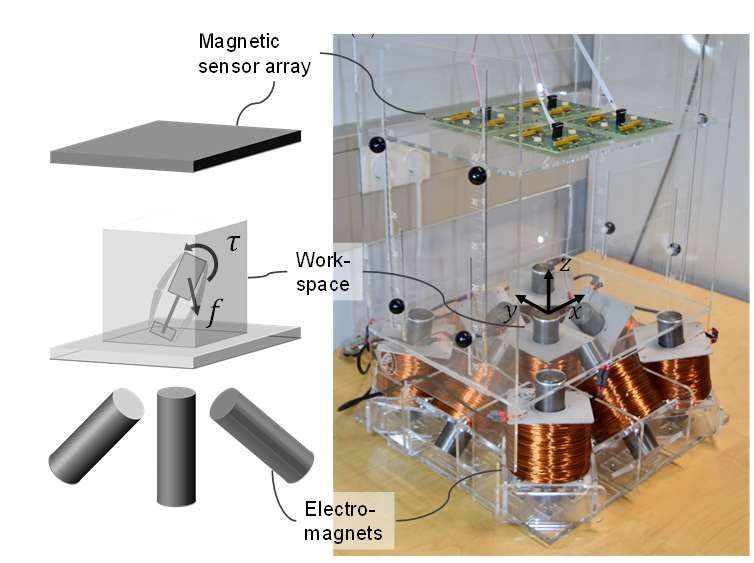}
	\caption{Actuation system of the MASCE}
	\label{fig:actuation_system}
\end{figure}

\begin{figure}[t!]
	\centering
	\includegraphics[width=\columnwidth]{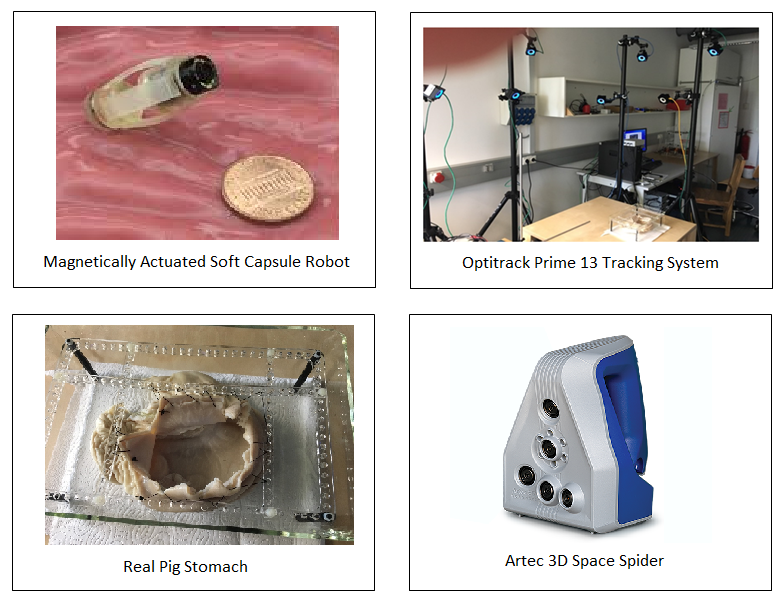}
	\caption{Experimental setup}
	\label{fig:exp_setup}
\end{figure}

\subsection{Visual Localization}
We propose the use of a direct surfel map fusion method for actively controllable endoscopic capsule robots. 
The core algorithm is the ElasticFusion method originally described by Whelan et al. \cite{whelan2016elasticfusion}, which uses a dense map and non-rigid model deformation to account for changing environments.
It performs joint volumetric and photometric alignment, frame-to-model predictive tracking, and dense model-to-model loop closure with non-rigid space deformation. 
Prior to using endoscopic video with such a method, the images must first be prepared.
\subsubsection{Multi-scale vessel enhancement and depth image creation}
\begin{figure}
	\centering
	\includegraphics[width=1\linewidth]{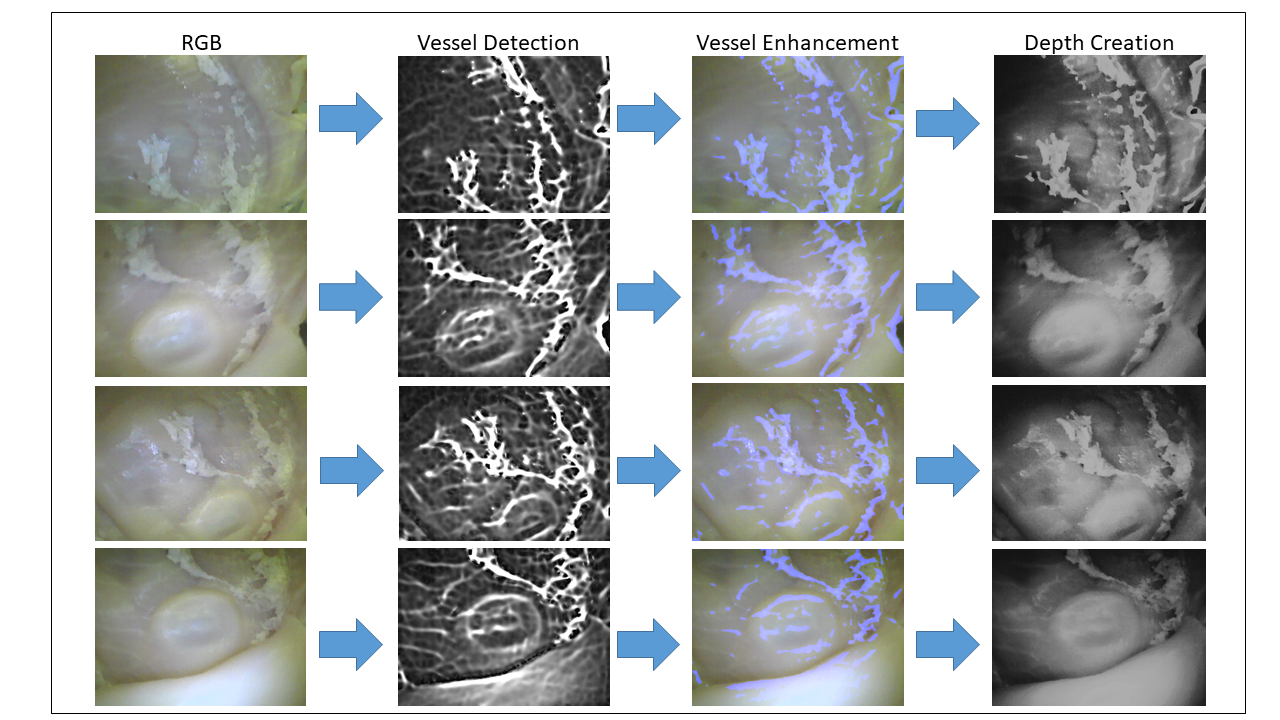}
	\caption{Input RGB image, vessel detection, vessel enhancement and depth image creation for sample frames.}
	\label{fig:datasam}
\end{figure}

Endoscopic images have mostly homogeneous and poorly textured areas. 
To prepare the camera frames for input into the ElasticFusion pipeline, our framework starts with a vessel enhancement operation inspired from \citep{frangi1998multiscale}. 
Our approach enhances blood vessels by analyzing the multiscale second order local structure of an image. First, we extract the Hessian matrix :
\begin{equation}
	H=\begin{bmatrix}
			I_{xx}& I_{xy}\\ 
			I_{yx}& I_{yy}
		\end{bmatrix}
		\end{equation}
where $I$ is the input image, and $I_{xx}$, $I_{xy}$, $I_{yx}$, $I_{yy}$ the second order derivatives, respectively. Secondly, eigenvalues $\left | \lambda_{1} \right |\leq  \left | \lambda_{2} \right |$ and principal directions $u_{1}$,  $u_{2}$ of the Hessian matrix are extracted. The eigenvalues and principal directions are then ordered and analyzed to decide whether the region belongs to a vessel. To identify vessels in different scales and sizes, multiple scales are created by convolving the input image and the final output is taken as the maximum of the vessel filtered image across all scales. For further details of the mathematical equations, the reader is referred to the original paper of \citep{frangi1998multiscale}. Figure \ref{fig:datasam} shows input RGB images, vessel detection and vessel enhancement results for four different frames.
 
To create the depth channel from input RGB data, we implemented a real-time version of the perspective shape from shading under realistic conditions \citep{pers} by reformulating the complex inverse problem for shading into a highly parallelized non-linear optimization problem, which we solve efficiently using GPU programming and a Gauss-Newton solver. 
Figure \ref{fig:datasam} shows samples of input RGB images and depth images created from them.

\subsubsection{Joint photometric-geometric pose estimation}

The vision-based localization system operates on the principle of optimizing both a photometric measure of error and a 3D geometric measure of error. 
The camera pose of the endoscopic capsule robot is described by a transformation matrix $\textbf{P}_t$:

\begin{equation}
\textbf{P}_t = \begin{bmatrix}
       \textbf{R}_t & \textbf{t}_t            \\[0.3em]
       0_{1\times 3} & 1           
     \end{bmatrix}
     \in \mathbb{SE}_3.
\end{equation}

Geometric pose estimation is performed by minimizing the energy cost function $E_{icp}$ between the current depth frame, $\mathcal{D}^l_t$, and the active depth model, $\hat{\mathcal{D}}_{t-1}^a$:
\begin{equation}
E_{icp} = \sum_k{((\textbf{v}^k - \exp{(\hat{\xi})}\textbf{T} \textbf{v}_k^t)\cdot\textbf{n}^k)^2}
\end{equation}
where $\textbf{v}^k_t$ is the back-projection of the $k$-th vertex in $\mathcal{D}^l_t$, $\textbf{v}^k$ and $\textbf{n}^k$ are the corresponding vertex and normal from the previous frame. $\textbf{T}$ is the estimated transformation from the previous to the current robot pose and $\exp{(\hat{\xi})}$ is the 
exponential mapping function from Lie algebra $\mathfrak{se}_3$ to Lie group $\mathbb{SE}_3$, which represents small changes The photometric pose $\xi$ between the current surfel-based reconstructed RGB image $\mathcal{C}^l_t$ and the active RGB model $\hat{\mathcal{C}}^a_{t-1}$ is determined by minimizing the photometric energy cost function: 
\begin{equation}
E_{rgb} = \sum_{\textbf{u} \in \Omega} \left( I(\textbf{u},\mathcal{C}^l_t) - I(\pi(\textbf{K} \exp(\hat{\xi}) \textbf{T} \textbf{p}(\textbf{u} , \mathcal{D}^l_t)), \hat{\mathcal{C}}_{t-1}^a) \right)^2
\end{equation}
where as above $\textbf{T}$ is the estimated transformation from previous to the current camera pose. 
The joint photometric-geometric pose optimization is defined by the cost function:
\begin{equation}
E_\textrm{track} = E_\textrm{icp} + w_\textrm{rgb}E_\textrm{rgb},
\end{equation}
with $w_\textrm{rgb} = 0.13$, which was determined experimentally to yield good performance for our datasets. 
For the minimization of this cost function in real-time, the Gauss-Newton method is employed. 
At each iteration of the method, the transformation $\textbf{T}$ is updated as $\textbf{T} \to \exp(\hat{\xi})\textbf{T}$.

\subsection{Particle Filtering based Magnetic-Visual Sensor Fusion}
\label{sec:fusion}

Particle filter is a statistical Bayesian filtering method that computes the posterior probability density function of sequentially obtained state vectors $\mathbf{x_{t}} \in \mathcal{X}$ which are suggested by sensor measurements.
We have implemented the method of Caron et al., which provides robustness against sensor failure through the introduction of latent variables characterizing the sensor's reliability as either normal or failing, which are estimated along with the system state \cite{caron2007particle}. 
The method is briefly described in what follows.

The state $\mathbf{x_{t}}$ composes the 6-DoF pose for the capsule robot, which is assumed to propagate in time according to a transition model:
\begin{equation}
\label{eq:state_transition}
\mathbf{x}_{t} = f(\mathbf{x}_{t-1} , \mathbf{v}_{t} )  
\end{equation}
where $f$ is a non-linear state transition function and $\mathbf{v_{t}}$ is a white noise. 
$t$ is the index of a time sequence, $t \in \{1,2,3, ...\}$. 

6-DoF pose state estimation with a high precision often requires multi-sensor input or sequential observations. 
The endoscopic capsule is equipped with two sensor systems, one being a 5-DoF magnetic sensor array and the other one being an endoscopic monocular RGB camera. 
Observations of the pose are produced by $n$ sensors $\mathbf{z}_{k,t}  (k = 1,...,n)$ in general,  where the probability distribution $p(\mathbf{z}_{k,t}|\mathbf{x}_t)$ is known for each sensor.

\begin{figure}
	\centering
	\includegraphics[width=0.9\linewidth]{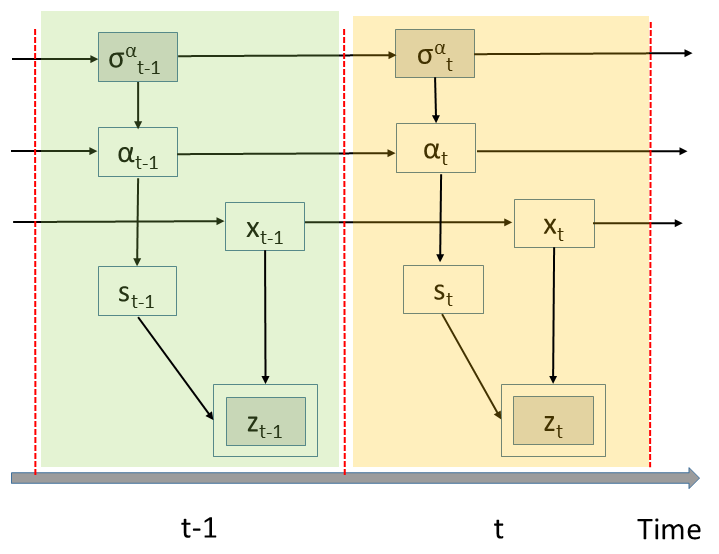}
	\caption{The switching state-space model. The double rectangles denote observable variables and the gray rectangles denote hyper-parameters.}
	\label{fig:graph_model}
\end{figure}

We estimate the 6-DoF pose states which rely on latent (hidden) variables by using the Bayesian filtering approach. 
The relations between all of the variables are shown in the probabilistic graphical model, Figure ~\ref{fig:graph_model}. The hidden variables of sensor states are denoted as $s_{k,t}$, which we call switch variables, where $s_{k,t}\in \{0,..., d_{k}\}$ for $k = 1,...,n$. $d_{k}$ is the number of possible observation models, e.g., failure and nominal sensor states. 
The observation model for $\mathbf{z}_{k,t}$ can be described as:

\begin{equation}
\mathbf{z}_{k,t} = h_{k, s_{k, t},t}(\mathbf{x}_{t}) + \mathbf{w}_{k, s_{k, t},t} 
\end{equation}
where $h_{k, s_{k, t},t}(\mathbf{x}_{t})$ is the non-linear observation function and $\mathbf{w}_{k, s_{k, t},t} $ is the observation noise. The latent variable of the switch parameter $s_{k, t}$ is defined to be $0$ if the sensor is in a failure state, which means that observation $\mathbf{z}_{k,t}$ is statistically independent of $\mathbf{x}_{t}$, and $1$ if the sensor $k$ is in its nominal state of work. The prior probability for the switch parameter $s_{k,t}$ being in a given state $j$, is denoted as  $\alpha_{k,j,t}$ and it is the probability for each sensor to be in a given state:

\begin{equation}
Pr(s_{k,t}=j) = \alpha_{k,j,t} , \quad 0 \leq j \leq d_{k} 
\end{equation}
where $\alpha_{k,j,t} \geq 0$ and  $\sum_{j=0}^{d_{k}}\alpha_{k,j,t}=1$ with a Markov evolution property. The objective posterior density function $p(\mathbf{x}_{0:t}, \mathbf{s}_{1:t}, \mathbf{\alpha}_{0:t}|\mathbf{z}_{1:t})$ and the marginal posterior probability $p(\mathbf{x}_{t}|\mathbf{z}_{1:t})$ , in general, cannot be determined in a closed form due to its complex shape. However, sequential Monte Carlo methods (\textit{particle filters}) provide a numerical approximation of the posterior density function with a set of samples (\textit{particles}) weighted by the kinematics and observation models.

\subsubsection{Sensor Failure Detection and Handling}

\begin{figure*}
\centering
\begin{subfigure}[t]{.47\textwidth}
\includegraphics[width=\textwidth]{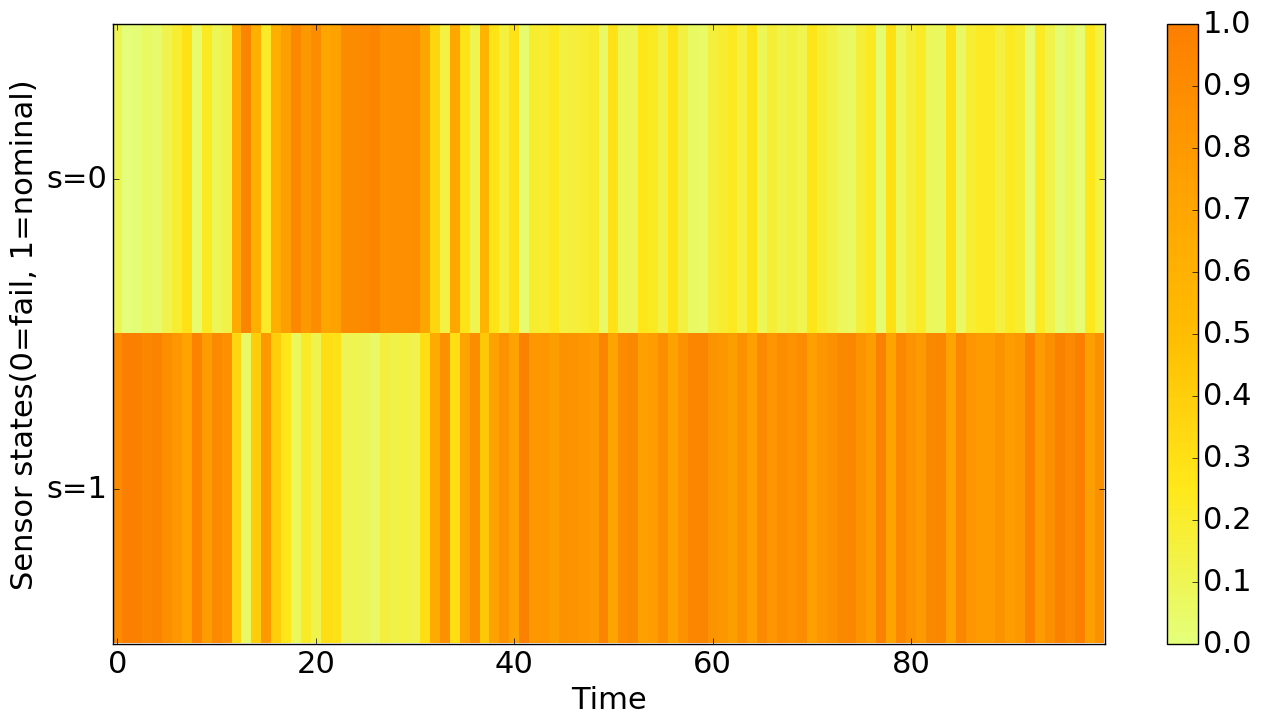}
%\caption{Translation loss decreases as the training proceeds.}
\label{fig:s1_result}       % Give a unique label
\end{subfigure} %
~ 
\begin{subfigure}[t]{0.47\textwidth}
\includegraphics[width=\textwidth]{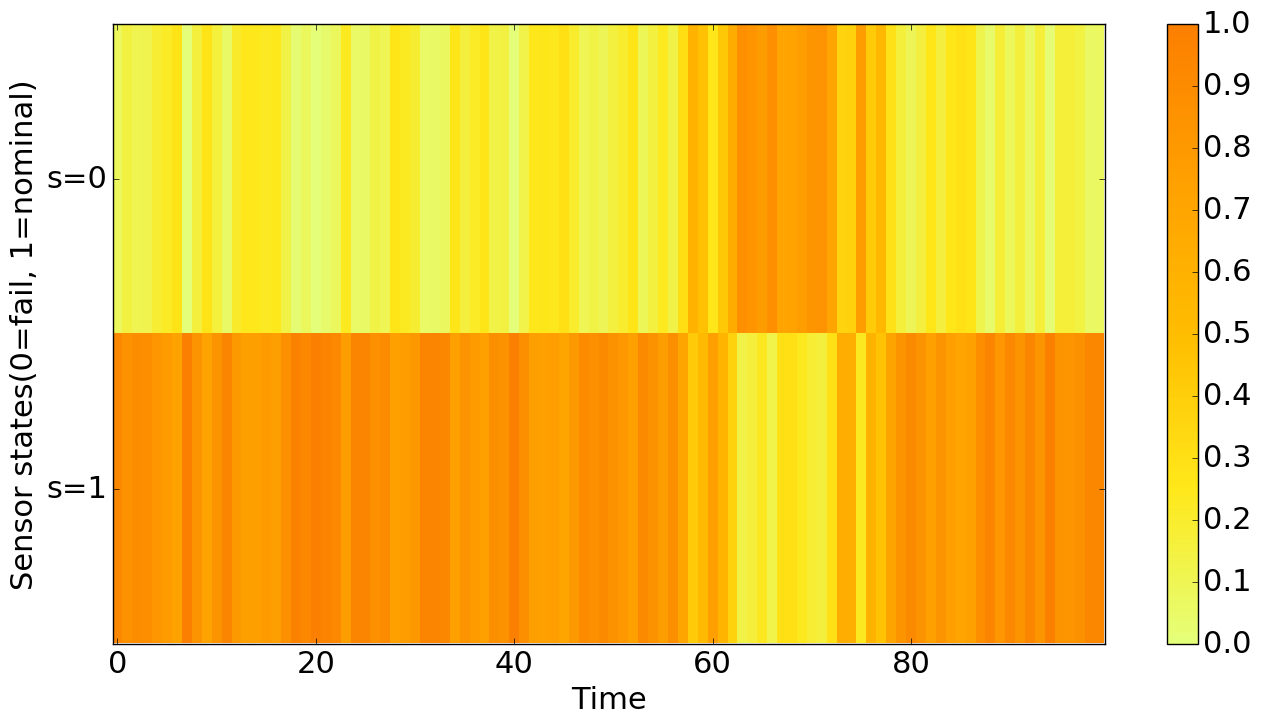}
%\caption{As the training proceeds, the rotation error decreases.}
\label{fig:s2_result}       % Give a unique label
\end{subfigure}

\begin{subfigure}[t]{.47\textwidth}
\includegraphics[width=\textwidth]{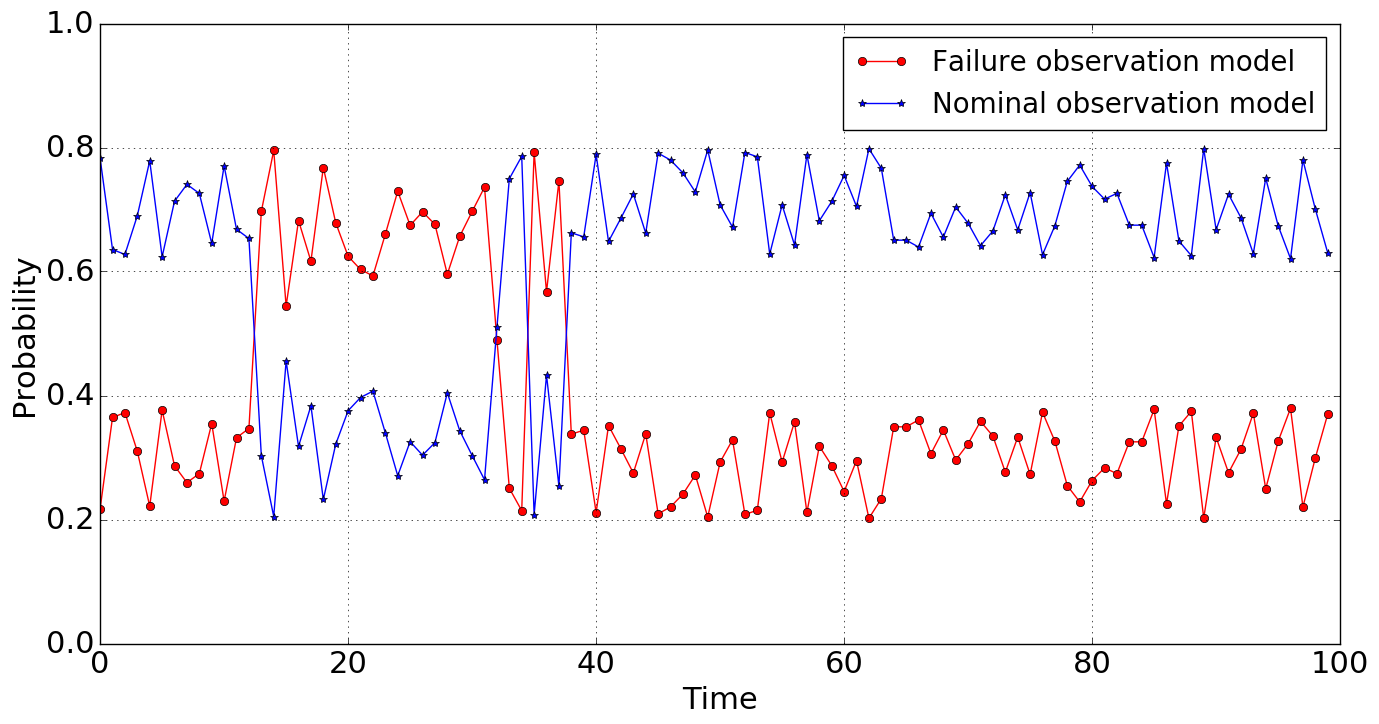}
%\caption{Translation loss decreases as the training proceeds.}
\label{fig:alpha1_result}       % Give a unique label
\end{subfigure} %
~ 
\begin{subfigure}[t]{0.47\textwidth}
\includegraphics[width=\textwidth]{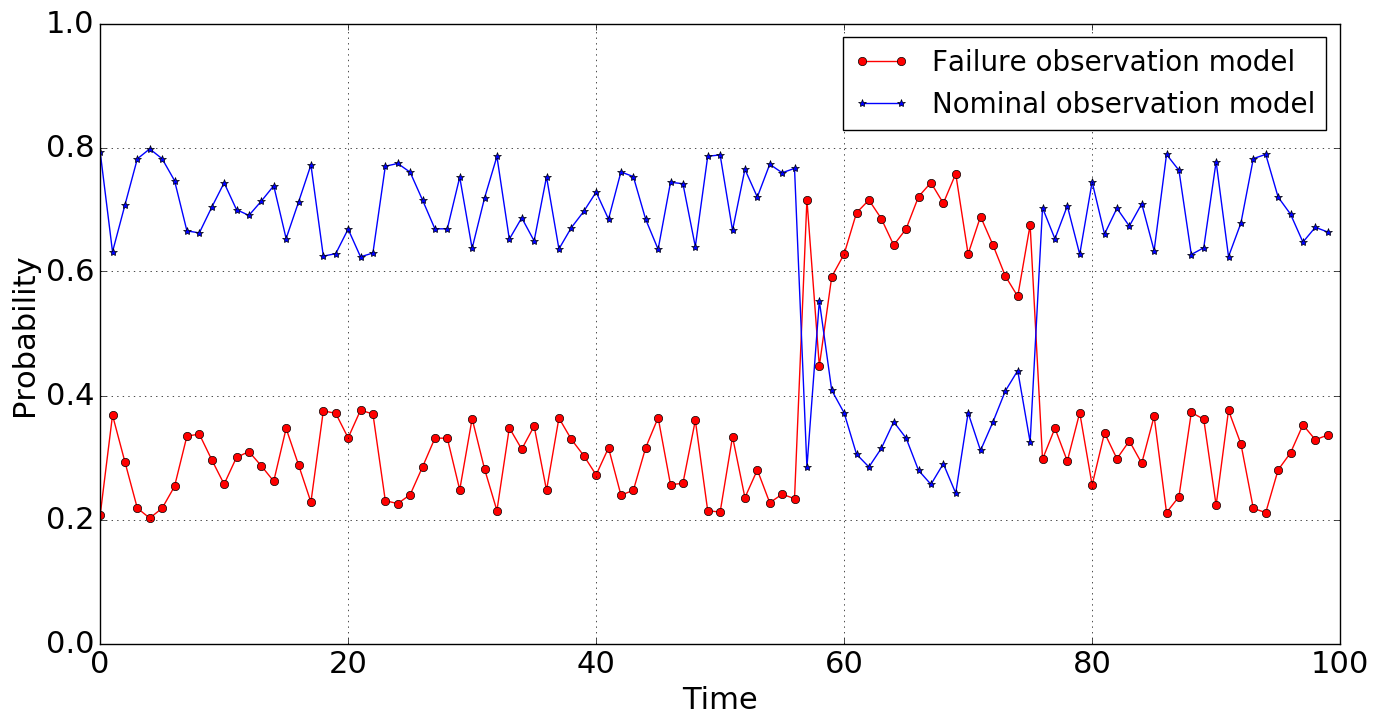}
%\caption{As the training proceeds, the rotation error decreases.}
\label{fig:alpha2_result}       % Give a unique label
\end{subfigure}

\caption{Top figures: Posterior probability of $\mathbf{s}_{k,t}$ parameter for endoscopic RGB camera (left) and for magnetic localization system (right). Bottom figures: The minimum mean square error (MMSE) of $\mathbf{\alpha}_{k,t}$ for endoscopic RGB camera (left) and for magnetic localization system (right).  The switch parameter, $\mathbf{s}_{k,t}$, and the confidence parameter $\mathbf{\alpha}_{k,t}$ reflect the failure times accurately: Visual localization fails between 14-36 seconds and magnetic sensor fails between 57-76 seconds. Both failures are detected confidentially.}
\label{fig:alpha_result}
\end{figure*}

\begin{figure}
\begin{subfigure}{0.47\columnwidth}
\includegraphics[width=\columnwidth]{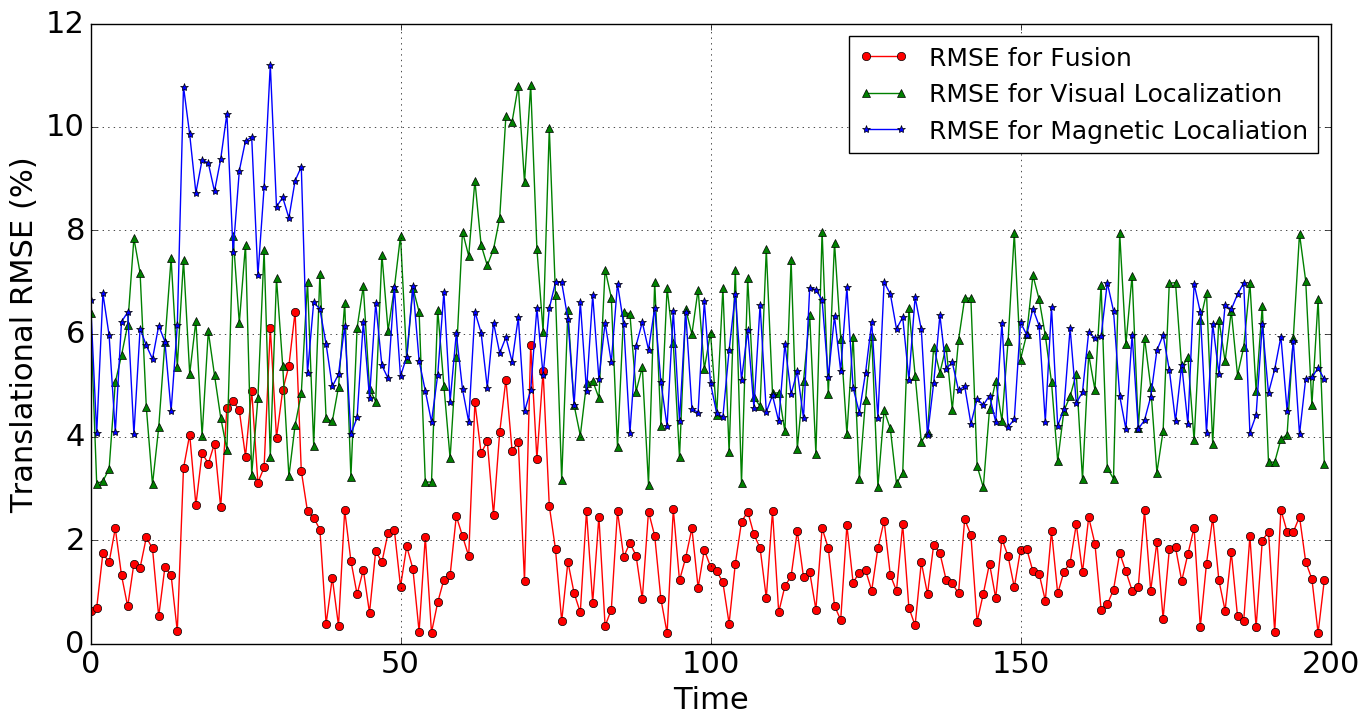}
%\caption{Translation loss decreases as the training proceeds.}
\label{fig:e1_result}       % Give a unique label
\end{subfigure}
~
\begin{subfigure}{0.47\columnwidth}
\includegraphics[width=\columnwidth]{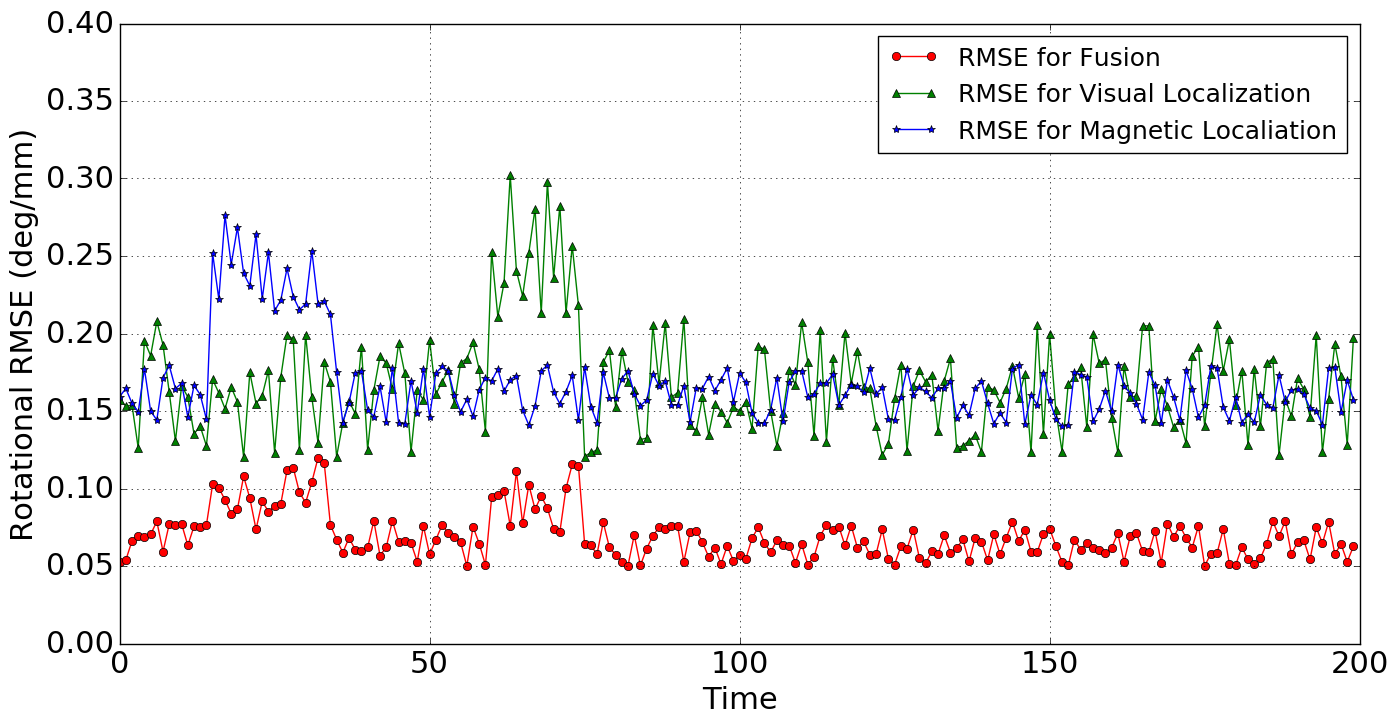}
%\caption{As the training proceeds, the rotation error decreases.}
\label{fig:e2_result}       % Give a unique label
\end{subfigure} 
\caption{Translational (top) and rotational (bottom) RMSEs for multi-sensor fusion, visual localization and magnetic localization.}
\label{fig:rmse_result}
\end{figure}

The proposed multi-sensor fusion approach is able to detect the sensor failure periods and to handle the failures, accordingly. As seen in Figure \ref{fig:alpha_result}), The posterior probabilities of the switch parameters $\mathbf{s}_{k, t}$ and the minimum mean square error (MMSE) estimates of $\mathbf{\alpha}_{k,t}$ indicate an accurate detection of sensor failure states. Visual localization failed between seconds 14-36 due to very fast frame-to-frame motions  and magnetic sensor failed between seconds 57-76 due to increased distance of the ringmagnet to the sensor array. Thanks to the switching option ability from one observation model to another in case of a sensor failure so that RMSE is kept low during sensor failure as seen in Figure \ref{fig:rmse_result}. In our sensor failure model, we do not make a Markovian assumption for the switch variable $\mathbf{s}_{k,t}$ but we do for its prior $\mathbf{\alpha}_{k,t}$, resulting in a priori dependent on the past trajectory sections, which is more likely for the incremental endoscopic capsule robot motions. 
The model thus introduces a memory over the past sensor states rather than simply considering the last state. The length of the memory is tuned by the hyper-parameters $\sigma^{\alpha}_{k,t}$,  leading to a long memory for large values and vice-versa. This is of particular interest when considering sensor failures. Our system detects automatically failure states. Thus, the confidence in the RGB sensor decrease when visual localization fails recently due to occlusions, fast-frame-to frame changes etc. On the other hand, the confidence in magnetic sensor decreases if the magnetic localization fails due to noise interferences from environment or if the ringmagnet has a big distance to the magnetic sensor array.

\subsubsection{LSTM-based Kinematic Model}
Existing sensor fusion methods based on traditional particle filter and Kalman filter approaches have limited functionality in non-linear dynamic systems. It is assumed that, in the Kalman filter and extended Kalman filter, the underlying dynamic process is well-modelled by linear equations or that these equations can be linearised without a major loss of fidelity. On the other hand, particle filters accommodate a wide variety of dynamic models, allowing for highly complex and non-linear dynamics in the state variables with a realistic non-linear kinematics. Unlike traditional feed-forward and just-in-moment artificial neural networks, LSTMs are very suitable for modelling the dependencies across time sequences and for creating a temporal motion model since it has a memory of hidden states over time and has directed cycles among hidden units, enabling the current hidden state to be a function of arbitrary sequences of inputs. Thus, using an LSTM, the pose estimation of the current time step benefits from information encapsulated in previous time sequences and is suitable to formulate the state transition function $f$ in Equation \ref{eq:state_transition}. In particular, information about the most recent velocities and accelerations become available to the model.

The training data is divided into input sequences of length $50$ and the slices are passed into the LSTM modules with the expectation that it predicts the next 6-DoF pose value, i.e. the $51\text{st}$ pose measurement, which was used to compute the cost function for training.
The LSTM module was trained using Keras library with GPU programming and Theano back-end. Using back-propagation-through-time method, the weights of hidden units were trained for up to $200$ epochs with an initial learning rate of $0.001$. Overfitting was prevented using dropout and early stopping techniques.

\subsection{Coupling of Magnetic and Vision Systems}
\label{sec:calib}

To couple the magnetic actuation and localization system which is seen in Fig. \ref{fig:actuation_system} with the proposed vision based SLAM system, an eye-in-hand calibration problem must be solved.
The vision system measures the pose of the camera, and the magnetic localization system measures the 5D pose of the magnet on the MASCE. 
The transformation between the coordinate frames attached at the ring magnet and at the camera origin must be known, because the particle filter assumes that the two systems make measurements on the same system state, which in this case is a single rigid body pose associated with the capsule.

There does not exist a standard method for determining the transformation matrix when one of the measurement systems is 5 DoF. 
We applied the standard theory for two 6 DoF sensors by assuming a value for the remaining rotational DoF in the magnetic sensor data. 
We then used the dual quaternion-based algorithm to determine the unknown transformation from the data \cite{daniilidis1999hand}. 

When the calibration is known, then the actuation and vision systems are automatically registered. 
This will facilitate the closing of the perception-action loop in future work by a planner and/or human teleoperator that uses the map to generate desired capsule trajectories in the magnetic coordinate system.

%For that purpose, we make use of dual quaternion based hand-eye calibration method proposed by \cite{daniilidis1999hand}, which relies on the invariance of the angle and pitch provided by the dual quaternion parametrization.  We denote by $\mathbf{X}$ the transformation from magnetic actuation system to the capsule camera, by $\mathbf{A}_{i}$ the transformation matrix between camera frame $i$ and frame $i+1$ , and by $\mathbf{B}_{i}$ the transformation matrix from magnetic actuation input $i$ to magnetic actuation input $i+1$. Figure \ref{fig:vis_mag_calib} illustrates the schema of our actuator-to-sensor transformation approach. For solving the transformation problem, we collect different capsule robot poses, which gives us different instances of the equation $\mathbf{A}\mathbf{X} = \mathbf{X}\mathbf{B}$ with the unknown transformation matrix $\mathbf{X}$. Using SVD (Singular Value Decomposition), we solve the equations and determine the actuator-to-camera transformation matrix $\mathbf{X}$. 

%\begin{equation} \label{eq:calib_trans}
%\mathbf{A}\mathbf{X} = \mathbf{X}\mathbf{B}
%\end{equation}

%\begin{figure}
%	\centering
%	\includegraphics[width=0.5\columnwidth]{figures/vis-mag-calib.png}
	% figure caption is below the figure
%	\caption{The transformations between different frames at pose $i$ and pose $i+1$.}
%	\label{fig:vis_mag_calib}       % Give a unique label
%\end{figure}

\subsection{Scene representation, Deformation graph and Loop Closure}
Due to strict real-time concerns of the approach, instead of using every pixel for reconstruction, the method uses surfels, where each surfel has a position, normal, color, weight, radius indicating the local surface area around a given point, initialization timestamp and last updated timestamp. 
The ElasticFusion method uses a deformation graph to non-rigidly deform the model during mapping. 
Non-rigid deformation information is embedded in the surfels that comprise the map, and loop closures are applied frequently. 
The importance of accounting for non-rigid deformation is critical in the soft environment of the GI tract, where the map may need to be updated not only due to previous error in measurement but also due to real changes in the geometry of the environment.

To ensure globally consistent surface reconstruction, our framework closes loops with the existing map as those areas are revisited by fusing reactivated parts into the active model area and simultaneously deactivating surfels which have not been seen in a period of time. 
Inactive surfels are not used for either frame tracking or map fusion until a loop closure happens which activates them again. 
This has the advantage of continuous frame-to-model and  model-to-model tracking which provides viewpoint-invariance. Moreover, frame-to-model tracking has benefits over frame-to-frame tracking once the capsule robot encounters drifts and needs to relocalize.
For a detailed description of the method, the reader is referred to the original source \cite{whelan2016elasticfusion}.

\section{EXPERIMENTS AND RESULTS} \label{sec:experiements}

We evaluate the performance of our system both quantitatively and qualitatively in terms of trajectory estimation, surface reconstruction accuracy and computational performance. 

\subsection{Dataset} \label{sec:dataset_equip}

Three different endoscopic cameras were used to capture the endoscopic videos; i.e AWAIBA Naneye, Misumi-V3506-2ES and POTENSIC camera. We mounted  endoscopic cameras on our magnetically activated soft capsule endoscope  (MASCE) systems. The dataset was recorded on an open real pig stomach. 
We scanned the stomach using the 3D Artec Space Spider image scanner. This 3D scan served as the ground truth for the evaluation of the 3D map reconstruction. 
To obtain the ground truth for 6-DoF camera pose, an OptiTrack motion tracking system consisting of eight infrared cameras and tracking software was utilized.
Figure \ref{fig:exp_setup} illustrates our experimental setup. 
A total of 15 minutes of stomach video was recorded containing over 10000 frames. 
Some sample frames of the dataset are shown in Figure \ref{fig:datasam}.

\subsection{Trajectory Estimation} \label{sec:trajectory_estimation}

\begin{figure}
	\begin{subfigure}{0.5\textwidth} 
		\includegraphics[width=\textwidth]{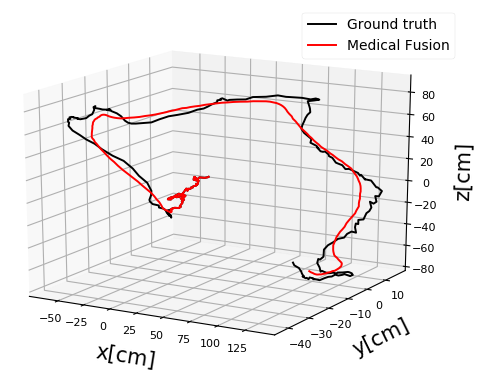}
		\caption{Trajectory 1}
		\label{fig:traj_1}       % Give a unique label
	\end{subfigure}
	\begin{subfigure}{0.5\textwidth} 
		\includegraphics[width=\textwidth]{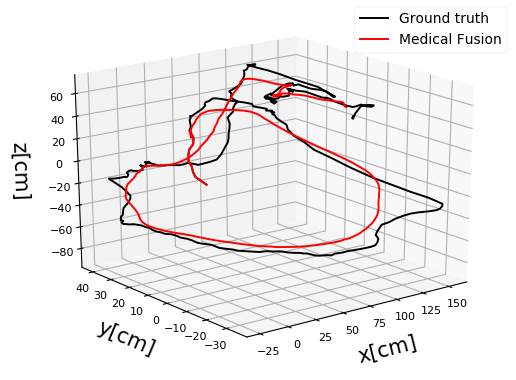}
		\caption{Trajectory 2}
		\label{fig:traj_2} 
	\end{subfigure}
	%\\
	\begin{subfigure}{0.5\textwidth} 
		\includegraphics[width=\textwidth]{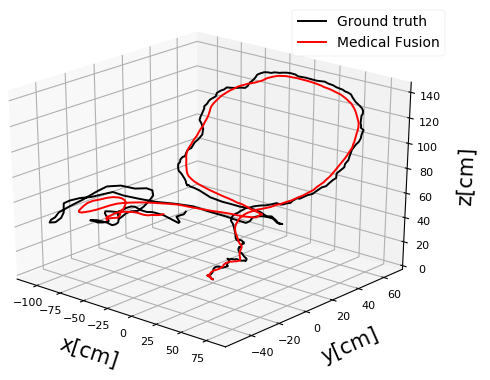}
		\caption{Trajectory 3}
		\label{fig:traj_3} 
	\end{subfigure}
	%~
	\begin{subfigure}{0.5\textwidth} 
		\includegraphics[width=\textwidth]{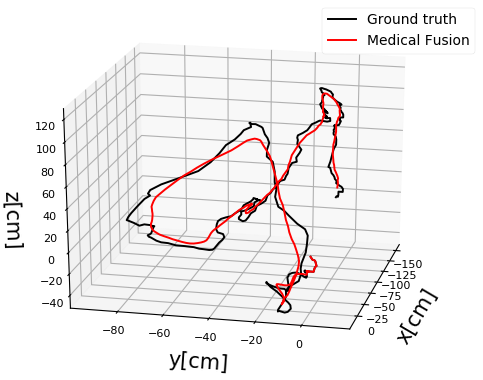}
		\caption{Trajectory 4}
		\label{fig:traj_4} 
	\end{subfigure}
	%~
	\caption{Sample ground truth trajectories and estimated trajectories.}
	\label{fig:trajectories}
\end{figure}

Table \ref{tab:trans_rmse} demonstrates the results of the trajectory estimation for four different trajectories that present varying levels of difficulty for the system. 
Trajectory 1 is an uncomplicated path with slow translations and rotations. Trajectory 2 follows a comprehensive scan of the stomach with many local loop closures. 
Trajectory 3 contains an extensive scan of the stomach with more complicated local loop closures. 
Trajectory 4 consists of more challenging motions including faster rotational and translational movements. 
Some qualitative tracking results of our proposed system and corresponding ground truth trajectories are demonstrated in Figure \ref{fig:trajectories}. 
For the quantitative analysis, we used the absolute trajectory (ATE) root-mean-square error between estimated pose values and the ground truth acquired by Optitrack system. 
As seen in Table \ref{tab:trans_rmse}, the system provides robust and accurate tracking in all of the tested conditions, and is not adversely affected by sudden movements, blur, noise or spectral reflections.
For the duration of the dataset, the Bayesian filtering algorithm estimated that the camera remained in the normally functioning sensor state at all times.

\begin{table}
	\centering
	% table caption is above the table
	\caption{Trajectory lengths and RMSE results in centimeters}
	\label{tab:trans_rmse}       % Give a unique label
	% For LaTeX tables use
	\begin{tabular}{ccccc}
		\hline\noalign{\smallskip}
		 Trajectory ID & POTENSIC & AWAIBA & MISUMI& LENGTH \\
		\hline\noalign{\smallskip}
		1&0.51&0.73&0.42 &41.4 \\
		\hline\noalign{\smallskip}
		2&0.64&0.81&0.48& 51.3 \\
		\hline\noalign{\smallskip}
		3&1.77&1.86&1.62& 43.2 \\
		\hline\noalign{\smallskip}
		4&1.92&1.95&1.73& 47.8 \\
		\hline\noalign{\smallskip}
	\end{tabular}
\end{table}

\subsection{Surface Reconstruction} \label{sec:surface_estimation}

\begin{table}
	\centering
	% table caption is above the table
	\caption{Absolute depth and 3D surface reconstruction RMSE results in centimeters}
	\label{tab:3drmse}       % Give a unique label
	% For LaTeX tables use
	\begin{tabular}{ccccc}
		\hline\noalign{\smallskip}
		Trajectory ID  & POTENSIC & MISUMI & AWAIBA& Length \\
		\hline\noalign{\smallskip}
		1 & 1.23 &1.25&1.28& 31.4 \\
		\hline\noalign{\smallskip}
		2 & 1.25 &1.27&1.35& 41.3 \\
		\hline\noalign{\smallskip}
		3 & 2.22 &2.26&2.37& 53.2 \\
		\hline\noalign{\smallskip}
		4 & 2.29 &2.31&2.39& 47.8 \\
		\hline\noalign{\smallskip}
	\end{tabular}
\end{table}

\begin{figure}
\centering
% Use the relevant command to insert your figure file.
% For example, with the graphicx package use
  \includegraphics[width=\textwidth]{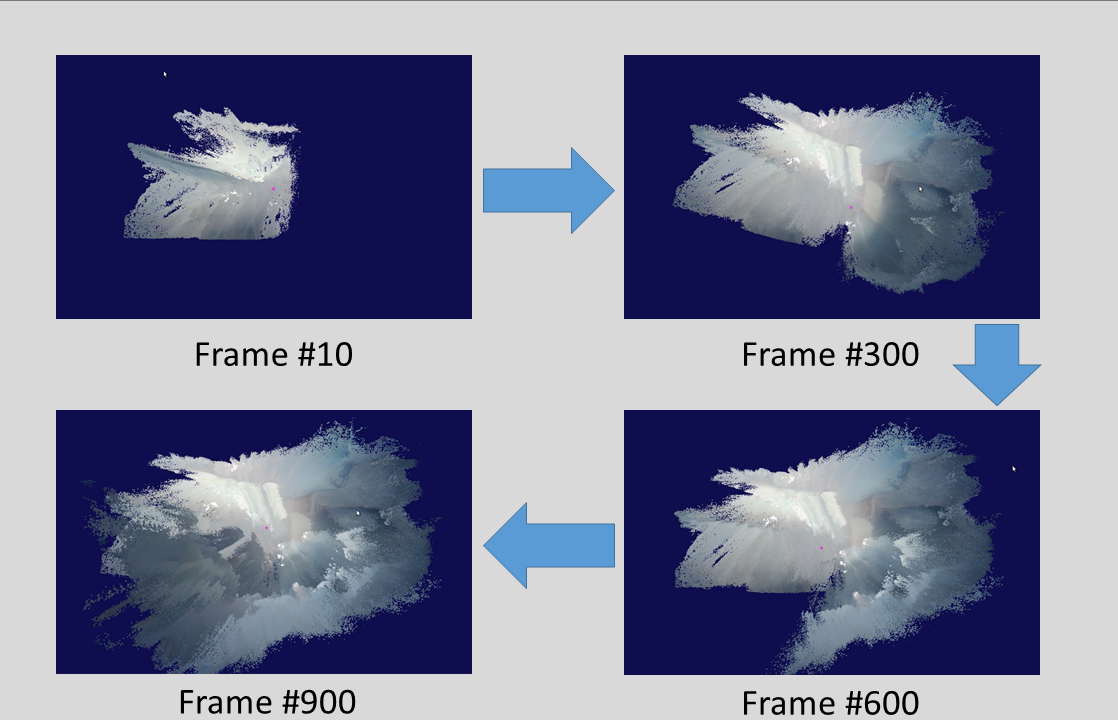}
% figure caption is below the figure
\caption{3D reconstructed soft stomach simulator surface}
\label{fig:recounstrudted_map}       % Give a unique label
\end{figure}

We scanned the non-rigid five different real porcine stomachs to obtain the ground truth for 3D map reconstruction. 
Reconstructed 3D surface and ground truth scan data were aligned using iterative closest point algorithm (ICP) and map reconstruction error is calculated using the absolute trajectory (ATE) RMSE between estimated depth values and the depth values of the scan data. 
We evaluated the surface reconstruction on the same four trajectories. 
The detailed RMSE results in Table \ref{tab:3drmse} prove that our system reconstructs 3D organ surface precisely. 
A sample 3D reconstruction process is shown in Figure \ref{fig:recounstrudted_map} for visual reference.

\subsection{Computational Performance} \label{sec:computational_performance} 

To analyze the computational performance of the system, we observed the average frame processing time across the sequences of Trajectory 1 to 4. 
The test platform was a desktop PC with an Intel Xeon E5-1660v3-CPU at 3.00 GHz, 8 cores, 32GB of RAM and an NVIDIA Quadro K1200 GPU with 4GB of memory. 
The execution time of the system is depended on the number of surfels in the map, with an overall average of 58 ms per frame scaling to a peak average of 67 ms implying a worst case processing frequency of 15 Hz.

\section{CONCLUSION} \label{sec:conclusion}

In this paper, we have presented a magnetic-RGB Depth fusion based SLAM method for endoscopic capsule robots. 
Our system makes use of surfel-based dense reconstruction in combination with particle filter based fusion of magnetic and visual localization information and sensor failure detection. 
The proposed system is able to produce a highly accurate 3D map of the explored inner organ tissue and is able to stay close to the ground truth endoscopic capsule robot trajectory even for challenging robot trajectories.
In the future, \textit{in vivo} testing is required to validate the accuracy and robustness of the SLAM system in the challenging conditions of the GI tract.
We also intend to extend our work into stereo capsule endoscopy applications to achieve even more accurate localization and mapping.
In addition, the inclusion of more sophisticated methods for the hand-eye calibration may result in improved accuracy.

\section*{References}

\bibliography{mybibfile}

\end{document}